\documentclass{article}

\usepackage{arxiv}

\usepackage[utf8]{inputenc} 
\usepackage[T1]{fontenc}    
\usepackage{hyperref}       
\usepackage{url}            
\usepackage{booktabs}       
\usepackage{amsfonts}       
\usepackage{nicefrac}       
\usepackage{microtype}      
\usepackage{lipsum}		
\usepackage{subcaption}
\usepackage{graphicx}
\usepackage{amsmath}
\usepackage{natbib}
\usepackage{doi}
\usepackage{orcidlink}

\title{Persistent Patterns in Eye Movements: A Topological Approach to Emotion Recognition}

\author{{Arsha Niksa\orcidlink{0009-0009-9757-3254}}\\
	Department of Mathematical Sciences\\
	Sharif University of Technology\\
	\texttt{arshaniksaa@gmail.com} \\
    \And
    {Hooman Zare\orcidlink{0009-0003-0496-726X}}\\
    Department of Mathematical Sciences\\
    Sharif University of Technology\\
    \texttt{hooman.hkz@gmail.com}\\
    \And
    {Ali Shahrabi\orcidlink{0009-0000-1908-4257}}\\
    Department of Physics\\
    Sharif University of Technology\\
    \texttt{alishahrabi97@gmail.com}\\
    \And
    {Hanieh Hatami\orcidlink{0009-0000-0217-2218}} \\
	Department of Physics\\
	Sharif University of Technology\\
	\texttt{haniehatami7@gmail.com} \\
	\And
    {Mohammadreza Razvan} \\
	Department of Mathematical Sciences\\
	Sharif University of Technology\\
	\texttt{razvan@sharif.edu} \\
}

\renewcommand{\shorttitle}{Persistent Patterns in Eye Movements}

\hypersetup{
pdftitle={A template for the arxiv style},
pdfsubject={q-bio.NC, q-bio.QM},
pdfauthor={David S.~Hippocampus, Elias D.~Striatum},
pdfkeywords={First keyword, Second keyword, More},
}

\begin{document}
\maketitle

\begin{abstract}
We present a topological pipeline for automated multiclass emotion recognition from eye-tracking data. Delay embeddings of gaze trajectories are analyzed using persistent homology. From the resulting persistence diagrams, we extract shape-based features such as mean persistence, maximum persistence, and entropy. A random forest classifier trained on these features achieves up to $75.6\%$ accuracy on four emotion classes, which are the quadrants the Circumplex Model of Affect.

The results demonstrate that persistence diagram geometry effectively encodes discriminative gaze dynamics, suggesting a promising topological approach for affective computing and human behavior analysis.
\end{abstract}


\section{Introduction}
Eye movements carry information about underlying cognitive and emotional states \citep{VR}. For example, gaze direction influences facial emotion perception: direct gaze facilitates recognition of approach-oriented emotions (anger, joy) while averted gaze facilitates avoidance-oriented emotions (fear, sadness) \citep{adams_perceived_2003}. Eye movements not only reflect where attention is directed but also convey information about underlying cognitive and emotional processes \citep{adams2003effects,adams2005effects}. Eye-tracking provides a window into a user’s visual and affective processes \citep{VR}. 

Gaze metrics such as fixations, saccades and pupil size have thus been used as affective cues in many studies \citep{kaiser2020effects,frontiers2021emotional}. In emotion recognition research, common practice is to extract statistical features from gaze trajectories (e.g., fixation duration, saccade amplitude, pupil diameter) \citep{Tar}. For instance, Tarnowski et al. computed 18 eye movement features and achieved approximately 80\% classification accuracy (three emotion classes, SVM) \citep{Tar}. Such handcrafted features can perform well \citep{pubmed2014eye,pmc2022eye}, but they may fail to capture complex nonlinear patterns and temporal dynamics in the gaze data. An example of why these methods might generaly fail to capture nonlinearities are discussed in the following article \citep{Rouh}

Topological data analysis (TDA) offers a different perspective by examining the shape of data across scales \citep{ichinomiya_machine_2025}. Persistent homology, a key TDA method, identifies features such as connected components and loops that persist over a range of scales \citep{carlsson2009topology}. A standard approach for time series is to form a point cloud via time-delay (Takens) embedding and then compute the persistence diagram of that cloud. Delay embeddings recover the geometry of the underlying dynamical attractor \citep{Chung}, and the resulting topological features are invariant to the speed of traversal and robust to noise. This method has been successfully applied to various domains, including time series analysis and spatial data clustering \citep{pereira2015persistent}.

In this study, we apply these ideas to eye-gaze trajectories for emotion classification. We model the gaze as a 2D time series and embed it in higher-dimensional space using a sliding-window delay embedding (inspired by Takens’ theorem \citep{Chung}). We then compute Vietoris–Rips persistence diagrams (homology dimensions 0 and 1) of the embedded point cloud. Each diagram is vectorized by calculating summary statistics of its persistence values (for example mean, max, standard deviation, skewness, kurtosis, quartiles, and persistent entropy; though not all are used in our methodology) \citep{Mishra}. These persistence statistics form a fixed-length feature vector representing the topological structure of the eye movement. 

We evaluate our method on the VREED (VR Eyes: Emotions) dataset \citep{VR}. VREED contains eye-tracking (and physiological) data from 34 participants exposed to 12 immersive 360° video environments, selected to span the four quadrants of the circumplex model of affect \citep{VR}. Participants reported valence and arousal using the Self-Assessment Manikin (SAM) and discrete emotions via a visual-analog scale, providing labels for supervised learning. 

Our results show that the proposed topological features improve emotion classification accuracy compared to conventional gaze features. This suggests that persistent homology can capture subtle spatiotemporal patterns in gaze behavior linked to affect. The remainder of the paper is organized as follows. Section 2 describes the VREED dataset and its emotion labels. Section 3 details the methodology (delay embedding, persistent homology, feature extraction). Section 4 presents classification experiments and performance analysis. Section 5 discusses the implications and outlines future work. Code and supplementary materials are provided in Section 6

Please note that OpenAI's ChatGPT has been used to assist in editing the English text of this paper and in generating parts of the code for feature engineering and classification.

\section{Data}
\label{sec:data}
We utilized the VREED dataset, a multimodal affective dataset designed for emotion recognition in immersive virtual reality (VR) environments \citep{VR}. VREED includes behavioral (eye-tracking) and physiological (ECG and GSR) recordings, as well as self-reported affective states from 34 participants exposed to 12 emotion-eliciting 360° virtual environments (360-VEs). These VEs were pre-selected through a pilot study and mapped to quadrants of the Circumplex Model of Affect (CMA) \citep{CMA}. Participants reported their valence-arousal states using the Self-Assessment Manikin (SAM), discrete emotions using a Visual Analog Scale (VAS), and presence levels via validated presence questionnaires.

The eye-tracking data, sampled at 60 Hz, include gaze coordinates and blink indicators for both eyes, pre-processed and available in both raw and feature-extracted forms. Physiological signals were sampled at high frequencies (2000 Hz for GSR and 1000 Hz for ECG) and aligned with emotional labels corresponding to each stimulus. Extracted feature datasets contain 312 labeled samples for both modalities. Labels reflect four emotional states based on the valence-arousal plane: High Arousal High Valence, Low Arousal High Valence, Low Arousal Low Valence, and High Arousal Low Valence.

As shown in Table~\ref{tab:emotion_freqs}, the dataset exhibits a high degree of class balance across the four emotional quadrants defined by the Circumplex Model of Affect. Each quadrant contains approximately 99–102 samples, indicating strong class homogeneity. This minimizes the risk of classifier bias due to class imbalance and enables more reliable evaluation of model performance across the entire emotional spectrum.

\begin{table}[h]
	\caption{Emotion Class Distribution in VREED Eye-Tracking Data}
	\centering
	\begin{tabular}{llr}
		\toprule
		Class Label & Emotion Quadrant (Valence/Arousal) & Frequency \\
		\midrule
		0 & High Valence, High Arousal     & 99 \\
		1 & High Valence, Low Arousal      & 99 \\
		2 & Low Valence, Low Arousal       & 102 \\
		3 & Low Valence, High Arousal      & 99 \\
		\bottomrule
	\end{tabular}
	\label{tab:emotion_freqs}
\end{table}

Each subject was exposed to 12 videos, each designed to elicit a distinct emotional response. For the purposes of this analysis, we focused solely on the gaze trajectory of the left eye. All trajectories were concatenated into a single dataset and shuffled, resulting in a total of 399 time series samples. To mitigate computational costs in subsequent topological computations, we downsampled each trajectory by selecting every 20\textsuperscript{th} index.

For evaluation, we adopted a three-way split of the dataset. First, 20\% of the data was held out as a final validation set. The remaining 80\% was further divided into training and testing sets with an 80/20 split, respectively. This setup ensures that model performance can be reliably assessed while retaining enough data for learning robust features.

\section{Methodologies}
\label{sec:methods}

\subsection{Topological Representation of Gaze Trajectories}
Let \( \mathcal{T} = \{ (x_t, y_t) \in \mathbb{R}^2 \mid t = 1, \dots, N \} \) denote a discrete eye-gaze trajectory. We equip \( \mathcal{T} \) with the Euclidean metric \( \|\cdot\|_2 \). A \textit{filtration} \( \{K_\epsilon\}_{\epsilon \geq 0} \) is a nested family of simplicial complexes indexed by a scale parameter \( \epsilon \). For \( \mathcal{T} \), we construct the \textit{Vietoris–Rips complex} \( \mathrm{VR}_\epsilon(\mathcal{T}) \), where a \( k \)-simplex spans \( k+1 \) points in \( \mathcal{T} \) with pairwise distances \( \leq \epsilon \).  

\begin{figure}[h]
    \centering
    \includegraphics[width=0.5\textwidth]{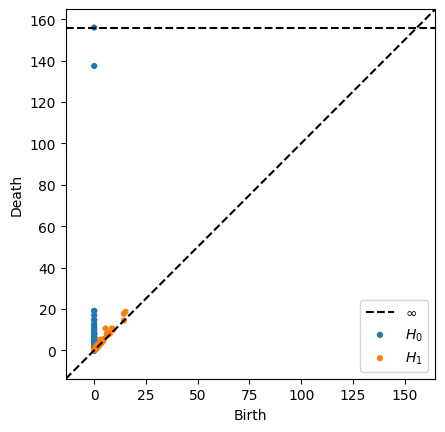}
    \caption{Persistence diagram of a subject's gaze trajectory. Each point corresponds to a topological feature, with birth time on the horizontal axis and death time on the vertical axis. Features near the diagonal have shorter lifespans and are more likely to be noise. This diagram was generated using the Ripser library for computing persistent homology and Persim for visualization.}
    \label{fig:pers_traj}
\end{figure}

The \textit{persistent homology} of this filtration captures topological features across scales. For each homological dimension \( p \in \mathbb{N}_0 \), we consider the persistence module \( \{H_p(K_\epsilon)\}_{\epsilon \geq 0} \) and its decomposition:
\[
\bigoplus_{i \in I} \mathbb{I}[b_i, d_i),
\]
where \( \mathbb{I}[b_i, d_i) \) is an interval module corresponding to a \( p \)-dimensional topological feature (e.g., connected components for \( p=0 \), loops for \( p=1 \)) born at scale \( b_i \) and dying at \( d_i \). The multiset of intervals \( \{(b_i, d_i)\}_{i \in I} \) is summarized in a \textit{persistence diagram} \( \mathcal{D}_p \subset \overline{\mathbb{R}}^2 \), where \( \overline{\mathbb{R}} = \mathbb{R} \cup \{\infty\} \). The horizontal axis encodes the scale at which a feature appears (birth), and the vertical axis shows when it disappears (death). Features that lie far from the diagonal are typically more meaningful, as they persist over longer ranges of the filtration. An example persistence diagram computed from the gaze trajectory of a subject is shown in Figure~\ref{fig:pers_traj}. Readers seeking a formal introduction to these concepts are referred to \citep{edelsbrunner2010computational}.

For computational efficiency, we downsample $\mathcal{T}$ by reduction factor $r$:
\[
\mathcal{T}_r = \{(x_{t_k}, y_{t_k}) \mid t_k = k \cdot r,  k = 1, \dots, \lfloor N/r \rfloor \}
\]
where $r=20$ balances temporal resolution and computational complexity.

\subsection{State Space Reconstruction via Takens Embedding}
We reconstruct the state space dynamics of gaze trajectories using Takens' embedding theorem. For each coordinate time series \( \{x_t\}_{t=1}^N \) and \( \{y_t\}_{t=1}^N \), define delay embeddings \( \phi_x, \phi_y: \mathbb{R} \to \mathbb{R}^d \) with embedding dimension \( d \) and delay \( \tau \):
\[
\phi_x(t) = \left( x_t, x_{t+\tau}, \dots, x_{t+(d-1)\tau} \right), \quad 
\phi_y(t) = \left( y_t, y_{t+\tau}, \dots, y_{t+(d-1)\tau} \right).
\]
The full state space embedding is then \( \Phi: \mathbb{R} \to \mathbb{R}^{2d} \):
\[
\Phi(t) = \left( \phi_x(t), \phi_y(t) \right).
\]
Parameters \( (d, \tau) \) are usually optimized per subject via mutual information minimization for \( \tau \) and false nearest neighbors analysis for \( d \) \citep{kantz2004nonlinear}. In this paper, however, we fixed parameters $d=3$, $\tau=10$, which were determined empirically. This yields three point clouds:
\begin{align*}
\mathcal{P}_{\mathrm{raw}} &= \mathcal{T} \subset \mathbb{R}^2, \\
\mathcal{P}_x &= \{\phi_x(t) \mid t = 1, \dots, N - (d-1)\tau \} \subset \mathbb{R}^d, \\
\mathcal{P}_y &= \{\phi_y(t) \mid t = 1, \dots, N - (d-1)\tau \} \subset \mathbb{R}^d.
\end{align*}

\subsection{Persistent Homology Computation}
For each point cloud $\mathcal{P} \in \{\mathcal{P}_{\mathrm{raw}}, \mathcal{P}_x, \mathcal{P}_y\}$, we compute the Vietoris-Rips filtration $\{\mathrm{VR}_\epsilon(\mathcal{P})\}_{\epsilon \geq 0}$. The persistence module is:
\[
H_p(\mathrm{VR}_\epsilon) \xrightarrow{i_{\epsilon, \epsilon'}} H_p(\mathrm{VR}_{\epsilon'}), \quad \epsilon \leq \epsilon'
\]
with $p$-dimensional persistent homology groups decomposed as $\bigoplus_{i \in I} \mathbb{I}[b_i, d_i)$. We compute persistence diagrams $\mathcal{D}_0$ (connected components) and $\mathcal{D}_1$ (loops) using the Ripser package.

Readers seeking a formal introduction to these concepts are referred to \citep{edelsbrunner2010computational}.

\subsection{Persistence-Weighted Feature Extraction}
For each diagram $\mathcal{D}_p$ ($p \in \{0,1\}$), let $\{(b_i, d_i)\}_{i=1}^n$ be birth-death pairs with persistence $\pi_i = d_i - b_i$. Define the persistence measure $\mu = \sum_{i=1}^n \pi_i \delta_{(b_i,d_i)}$. We extract for $\alpha \in \{b, d, \pi\}$:

\begin{enumerate}
    \item \textit{Mean}: \[\mu^{(\alpha)}=\frac{1}{n} \sum_{i=1}^n \alpha_i \]
    
    \item \textit{Entropy}: \[H^{\alpha}=-\sum_{i=1}^n p_i \log p_i , p_i = \frac{\alpha_i}{\sum_j \alpha_j} \]

    \item \textit{Maximum}:\[\max_i{\alpha_i}\]

    \item \textit{Cardinality of the Set for $\alpha$}:
    \[\text{card}(\{\alpha_1, ..., \alpha_n\}) = n\]
\end{enumerate}
We apply persistent homology to the trajectory and compute the features above from the resulting diagrams. The resulting vector is similar to a concept referred to as persistence statistics, introduced in \citep{chung2021persistence}. 


\subsection{Classification Framework}
Let $\{(\mathbf{f}_i, y_i)\}_{i=1}^K$ be labeled samples with $y_i \in \mathcal{C} = \{0,1,2,3\}$ (emotional states). We implement a Random Forest classifier:
\[
\hat{y} = \mathrm{mode}\left( \{ h_k(\mathbf{f}) \}_{k=1}^{n_{\text{trees}}} \right)
\]
where $h_k$ are decision trees trained on bootstrap samples with feature subset size $\lfloor \sqrt{m} \rfloor$. Hyperparameters: $n_{\text{trees}}=100$, Gini impurity splitting.

\subsection{Computational Implementation}
Persistence diagrams were computed using Ripser.py \citep{tralie2018ripser}. The classification pipeline was implemented in scikit-learn \citep{scikit-learn}.

\section{Results}
\label{sec:results}

We evaluated our topological feature-based classification model (Random Forest) against a baseline model trained on processed features provided in the VREED dataset \citep{VR}. Both models were evaluated on a 4-class emotion recognition task using eye trajectory data.

\subsection*{Topological Features (Our Method)}

\begin{figure}[h]
    \centering
    \includegraphics[width=0.45\textwidth]{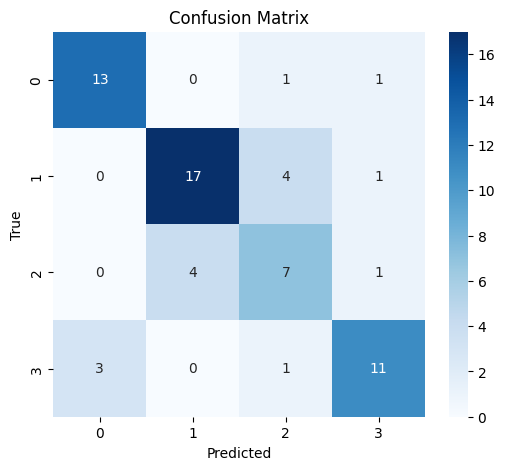}
    \caption{Confusion matrix for the test set using our method (topological features).}
    \label{fig:conf1}
\end{figure}

Our method achieved a test accuracy of 75.0\%, with a validation accuracy of 75.7\%. The performance across classes was relatively balanced, with F1-scores ranging from 0.56 to 0.84. Notably, classes 0 and 1 achieved the highest F1-scores (0.84 and 0.79, respectively), while class 2 was the most difficult to classify. The detailed classification report is shown in Table~\ref{tab:our_class_report}, and the corresponding confusion matrix is visualized in Figure~\ref{fig:conf1}.

\begin{table}[h]
\centering
\caption{Classification report for our method (Random Forest on topological features)}
\begin{tabular}{lcccc}
\toprule
Class & Precision & Recall & F1-score & Support \\
\midrule
0 & 0.81 & 0.87 & 0.84 & 15 \\
1 & 0.81 & 0.77 & 0.79 & 22 \\
2 & 0.54 & 0.58 & 0.56 & 12 \\
3 & 0.79 & 0.73 & 0.76 & 15 \\
\midrule
\textbf{Accuracy} & \multicolumn{4}{c}{\textbf{0.75}} \\
\bottomrule
\end{tabular}
\label{tab:our_class_report}
\end{table}

\subsection*{Baseline: Processed Features}

\begin{figure}[h]
    \centering
    \includegraphics[width=0.45\textwidth]{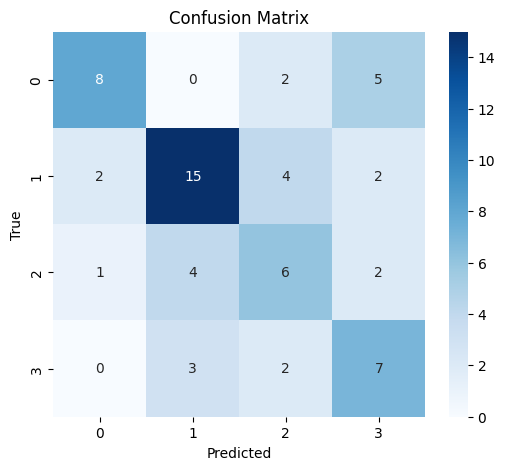}
    \caption{Confusion matrix for the baseline method.}
    \label{fig:conf2}
\end{figure}

The baseline model trained on standard processed features reached a lower accuracy of 57.1\%. Performance was notably weaker for classes 2 and 3, with F1-scores of 0.44 and 0.50, respectively. The full classification report is provided in Table~\ref{tab:baseline_class_report}, and the confusion matrix is shown in Figure~\ref{fig:conf2}.

\begin{table}[h]
\centering
\caption{Classification report for baseline method (processed features)}
\begin{tabular}{lcccc}
\toprule
Class & Precision & Recall & F1-score & Support \\
\midrule
0 & 0.73 & 0.53 & 0.62 & 15 \\
1 & 0.68 & 0.65 & 0.67 & 23 \\
2 & 0.43 & 0.46 & 0.44 & 13 \\
3 & 0.44 & 0.58 & 0.50 & 12 \\
\midrule
\textbf{Accuracy} & \multicolumn{4}{c}{\textbf{0.571}} \\
\bottomrule
\end{tabular}
\label{tab:baseline_class_report}
\end{table}

\subsection*{Findings}

The use of persistent homology-derived topological features significantly improved classification performance across all metrics. The average F1-score increased from 0.56 (baseline) to 0.74 (our method), with the biggest gains observed in classes with more distinctive trajectory dynamics. This demonstrates the utility of topological signal processing in emotion recognition from eye movement data.

\section{Discussion}
\label{sec:discussion}

\begin{figure}[h]
    \centering
    \begin{subfigure}{0.45\textwidth}
        \includegraphics[width=\linewidth]{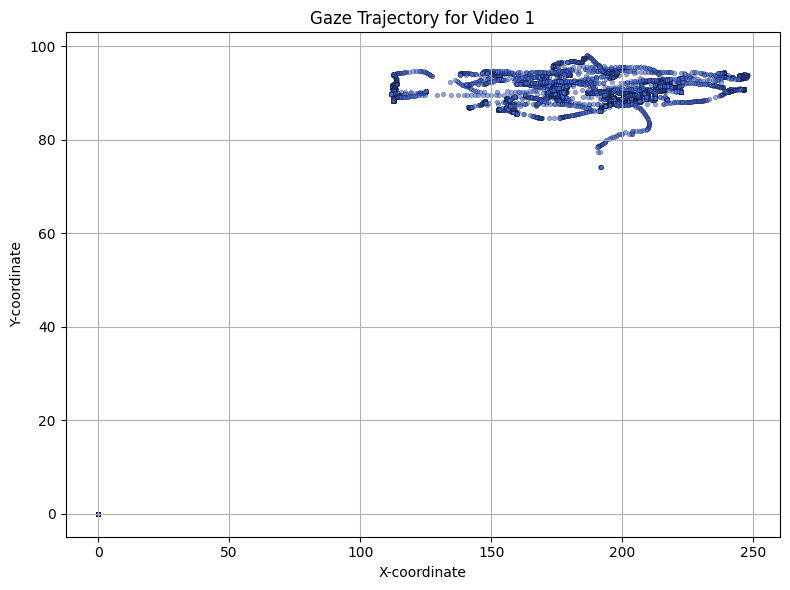}
        \caption{Class 0}
    \end{subfigure}
    \hfill
    \begin{subfigure}{0.45\textwidth}
        \includegraphics[width=\linewidth]{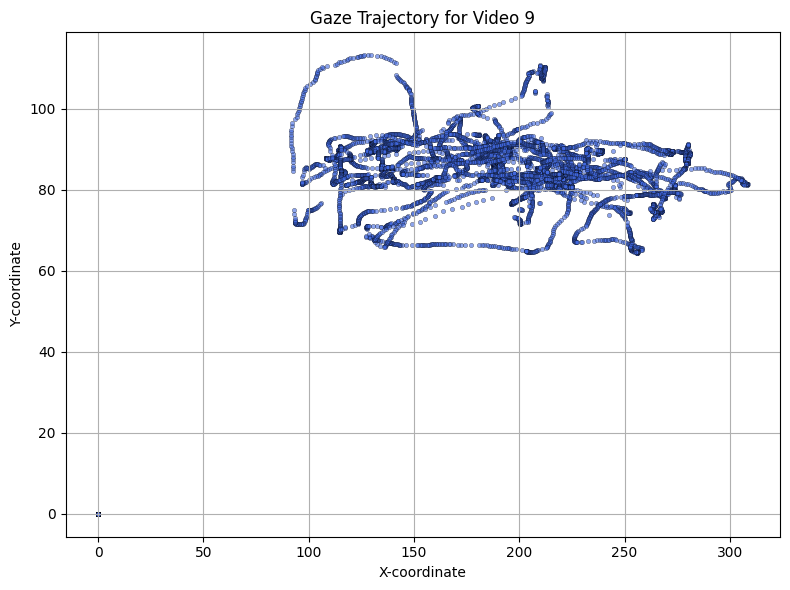}
        \caption{Class 1}
    \end{subfigure}
    \par\smallskip
    \begin{subfigure}{0.45\textwidth}
        \includegraphics[width=\linewidth]{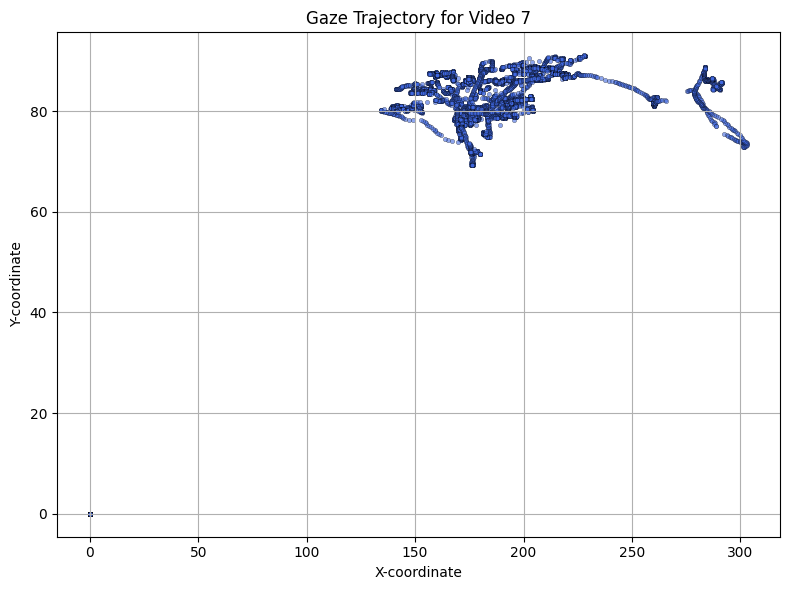}
        \caption{Class 2}
    \end{subfigure}
    \hfill
    \begin{subfigure}{0.45\textwidth}
        \includegraphics[width=\linewidth]{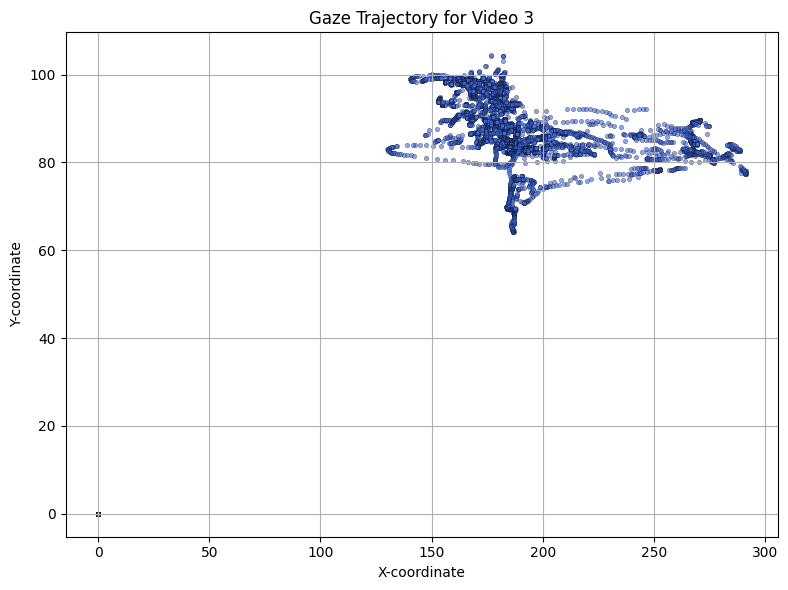}
        \caption{Class 3}
    \end{subfigure}
    \caption{Examples of gaze trajectories for each emotional quadrant. Each class exhibits a distinct geometric character, which may be captured effectively by topological features.}
    \label{fig:emo_shapes}
\end{figure}

Persistent homology provides a robust approach to extracting useful geometric and topological information from time series, and our results suggest that it is able to effectively discriminate between emotional states from eye movement records. Clearly, it must be possible for the different emotional quadrants to be distinguished by having different geometric patterns and consequently different homology groups. This is in accord with the comparatively different visual characters of each of the quadrants, as shown in Figure~\ref{fig:emo_shapes}.

The classification issue within this study is a toy problem. It seems there is potential to broaden the use of topological data analysis into many more fields of psychology and neurology. Persistent homology, for example, could be applied to the issues of ADHD diagnosis, default mode network analysis, and lie detection. Moreover, this hardware is not limited to gaze trajectories, it can be reused for face motion capture data, hand gestures, or even higher-dimensional representations of speech or text.

\begin{figure}[h]
    \centering
    \begin{subfigure}{0.45\textwidth}
        \includegraphics[width=\linewidth]{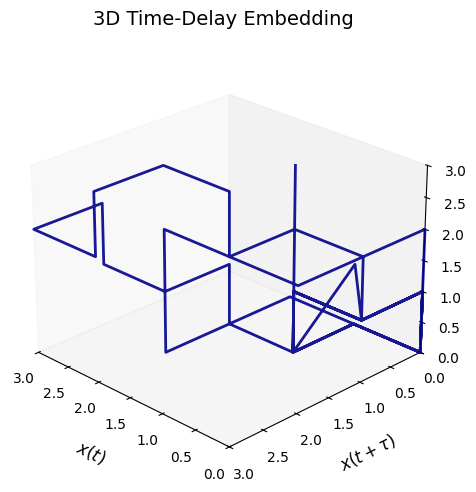}
        \caption{Example 1}
    \end{subfigure}
    \hfill
    \begin{subfigure}{0.45\textwidth}
        \includegraphics[width=\linewidth]{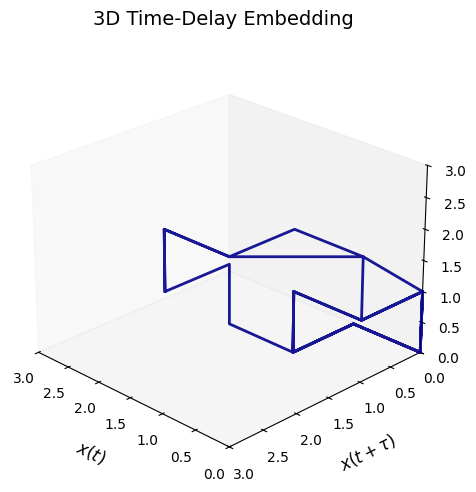}
        \caption{Example 2}
    \end{subfigure}
    \par\smallskip
    \begin{subfigure}{0.45\textwidth}
        \includegraphics[width=\linewidth]{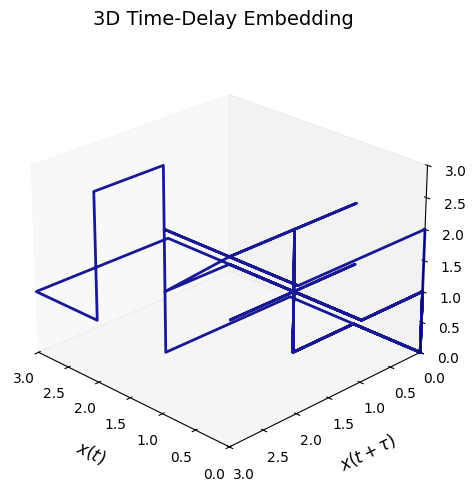}
        \caption{Example 3}
    \end{subfigure}
    \hfill
    \begin{subfigure}{0.45\textwidth}
        \includegraphics[width=\linewidth]{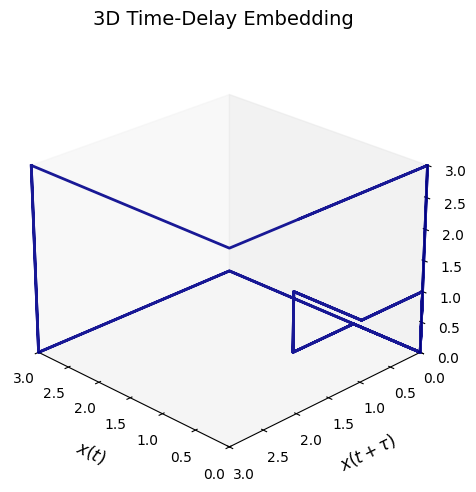}
        \caption{Example 4}
    \end{subfigure}
    \caption{Simplicial complexes derived from blink frequency embeddings. These geometric structures, while not yielding performance improvements here, may provide insights into the topological properties of discrete temporal phenomena.}
    \label{fig:simp_complexes}
\end{figure}

An intriguing theoretical direction is to use persistent homology on discrete event-based time series. One example is eye blinking: we can calculate the blink rate over time and embed the resulting sequences in higher-dimensional space (e.g., by using delay embeddings). By constructing simplicial complexes between adjacent points, we obtain ordered and often highly structured geometric objects. Such blink frequency embedding complexes are illustrated in Figure~\ref{fig:simp_complexes}. Although our original experiment did not discover that adding persistence features from blink information assisted in classification, this line of inquiry may be theoretically rich in importance for describing the topological structure of lean event-driven dynamics.

\section{Resources}
\label{sec:resources}

The code used for this study is available on GitHub and consists of two main Jupyter notebooks:

\begin{itemize}
    \item \textbf{PersistentHomology.ipynb} – Implements the proposed method using time-delay embeddings and persistent homology to extract features from eye movement trajectories.
    \item \textbf{NormalML.ipynb} – Implements a baseline machine learning method using conventional statistical features derived from the same dataset.
\end{itemize}

The notebooks are available at: \href{https://github.com/A-Niksa/gaze-persistent-homology}{GitHub Repository}.

\textbf{Note:} To execute the notebooks, users must download the dataset and correctly configure the data directory paths. The dataset used in this study is not included in the repository and should be downloaded separately. Please ensure that the directory structure matches the assumptions made in the notebooks before running the code.

\section*{Acknowledgements}

The authors would like to thank Kasra Khoshjoo for the valuable discussions and his technical expertise in feature engineering, which contributed meaningfully to the development of this project.

\bibliographystyle{unsrtnat}
\bibliography{references}  






\end{document}